\documentclass[11pt,a4paper]{article}
\usepackage[hyperref]{emnlp2020}
\usepackage{times}
\usepackage{latexsym}

\usepackage{booktabs}
\usepackage{graphicx}
\usepackage{amsmath}
\usepackage{enumitem}

\usepackage{todonotes}

\usepackage{microtype}

\aclfinalcopy 

\title{Evaluating Factuality in Generation with Dependency-level Entailment}

\author{Tanya Goyal \and Greg Durrett \\
  Department of Computer Science \\
  The University of Texas at Austin \\
  {\tt tanyagoyal@utexas.edu, gdurrett@cs.utexas.edu}}

\date{}

\begin{document}
\maketitle
\begin{abstract}
Despite significant progress in text generation models, a serious limitation is their tendency to produce text that is factually inconsistent with information in the input. Recent work has studied whether textual entailment systems can be used to identify factual errors; however, these sentence-level entailment models are trained to solve a different problem than generation filtering and they do not localize \emph{which part} of a generation is non-factual.
In this paper, we propose a new formulation of entailment that decomposes it at the level of dependency arcs. Rather than focusing on aggregate decisions, we instead ask whether the semantic relationship manifested by individual dependency arcs in the generated output is supported by the input. Human judgments on this task are difficult to obtain; we therefore propose a method to automatically create data based on existing entailment or paraphrase corpora. Experiments show that our dependency arc entailment model trained on this data can identify factual inconsistencies in paraphrasing and summarization better than sentence-level methods or those based on question generation, while additionally localizing the erroneous parts of the generation.
\footnote{Data and code available at \url{https://github.com/tagoyal/dae-factuality}}
\end{abstract}


\section{Introduction} 
The rise of pre-trained language models \cite{devlin2019bert, radford2019language} has led to strong text generation models for applications including summarization \cite{dong2019unified,Lewis2019BARTDS}, paraphrasing \cite{Goyal-Durrett:2020:Syntaxparaphrase, shen2020neural}, story generation \cite{mao2019improving}, and data augmentation \cite{wei2018fast, zhang2019addressing}. However, while these models generate fluent and grammatical text, they are prone to making factual errors that contradict the input text \cite{cao2018faithful}. Automatic metrics used to evaluate text generation, such as ROUGE and BERTScore \cite{Zhang2020BERTScore}, are not correlated with the factual consistency or faithfulness of the generated text \cite{falke2019ranking, kryscinski2019neural}. To address this, recent work has studied the use of textual entailment models to rank and filter non-factual generations \cite{falke2019ranking, maynez2020faithfulness}. However, these models suffer from issues such as dataset biases \cite{gururangan2018annotation,zhou2020towards} and a mismatch between the training data (entailment) and the test data (model generations). 

\begin{figure}
\centering
    \includegraphics[trim=70mm 60mm 50mm 35mm,scale=0.30,clip]{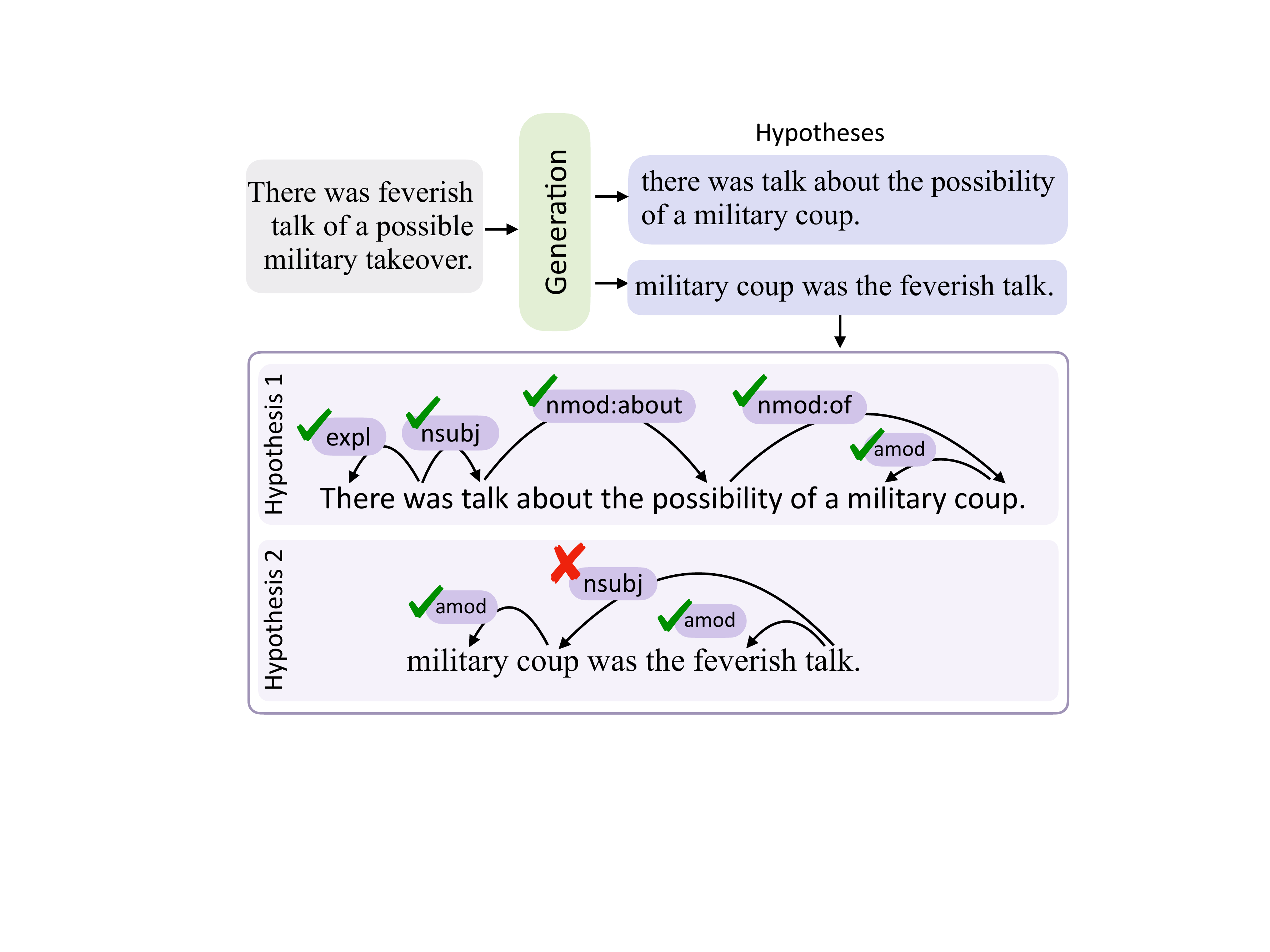}
    \caption{Overview of our dependency arc entailment formulation using a filtered set of Stanford Enhanced Dependencies. The DAE model makes independent factuality decisions for each dependency arc from the two generated hypotheses.} 
    \label{fig:overview} 
\end{figure} 

In this paper, we propose to decompose entailment decisions in a sentence to evaluate the faithfulness of generated text in a more fine-grained way. 
Rather than making a sentence-level entailment decision, we instead evaluate the entailment of dependency arcs of the generated sentence, as illustrated in Figure~\ref{fig:overview}. This approach views dependency arcs as semantic units that can be interpreted in isolation. Each arc is therefore judged \emph{independently} based on whether the relation it implies is entailed by the source sentence. This is helpful in localizing generation errors and consequently providing more interpretable model decisions. 


Decomposing the factuality evaluation over components of structured representations can also be extended to other formalisms like AMR \cite{banarescu2013abstract}, UDS \cite{white2016universal}, and more. The chief advantage of dependency parsing over these is that pre-existing tools for dependency parsing report very high performance. Another line of work focuses on question answering-based semantic representations \cite{fitzgerald2018large,michael2018crowdsourcing} or generating freeform questions to capture factuality \cite{wang2020asking, durmus2020feqa}. However, these systems require a separate question generation step at \emph{inference} time, so generation is baked into their formalisms in a heavyweight way whereas we \emph{only} require dependencies. A final advantage of our approach is that dependency arcs can be produced in an online fashion during inference, and hence, factuality can be enforced incrementally.

We evaluate our proposed dependency arc entailment approach in both summarization and paraphrase settings. In both settings, we show that we can automatically derive labels from actual generation data rather than rely on human annotation of dependency arc entailment, which is challenging to collect at scale. Nevertheless, our results show that our system's performance on factuality classification surpasses both sentence-level entailment and question generation and answering models.
Our derived labels from actual generation data provide much better task-specific supervision compared to general entailment datasets. Finally, we demonstrate that predicted entailment scores for individual dependency arcs are meaningful and can be leveraged to understand and localize errors in system generations.

\section{Dependency Arc Entailment (DAE)} 
\label{sec:DAE}



\paragraph{Defining arc entailment} Our notion of entailment starts by assuming a rough correspondence between predicates and arguments in two sentences. In natural language inference (NLI) annotation efforts, this has taken the form of anchoring judgments in an underlying imagined scene \cite{bowman-etal-2015-snli}. We make a similar assumption, that events and actors in the source and target sentences are in correspondence unless there is direct evidence to the contrary. For instance, in Figure~\ref{fig:overview}, the \emph{military coup} in the target sentence and its corresponding \emph{amod(coup$\rightarrow$military)} arc should be evaluated with respect to the \emph{military takeover} in the source, giving coreference of the two the benefit of the doubt here.

With this assumption, we say that a dependency arc in the target sentence is entailed by the source if the \emph{semantic} relationship it implies between its head and child is entailed by the source sentence. There is precedent for such a syntax-semantics correspondence: certain formalisms like meaning-text theory \cite{Melcuk1988} have historically made this mapping more or less explicit.
Consider the first hypothesis in Figure~\ref{fig:overview}. Many of the arcs here either contain information analogous to that in semantic roles, or they specify nominal modifiers capturing important entity properties.\footnote{We use enhanced dependencies in our experiments; modifiers with prepositions, augmented conjuncts provide a more useful semantic representation.} In our implementation, we exclude certain arc types which are not strongly tied to semantics, such as arcs involving punctuation; see the Appendix for details.
Note that our method does not support commenting on arcs of the input that do not exist in the output; we discuss this later in Section~\ref{sec:limitations}.


\begin{figure*}[t!]
\centering
    \includegraphics[trim=110mm 156mm 50mm 57mm,scale=0.35,clip]{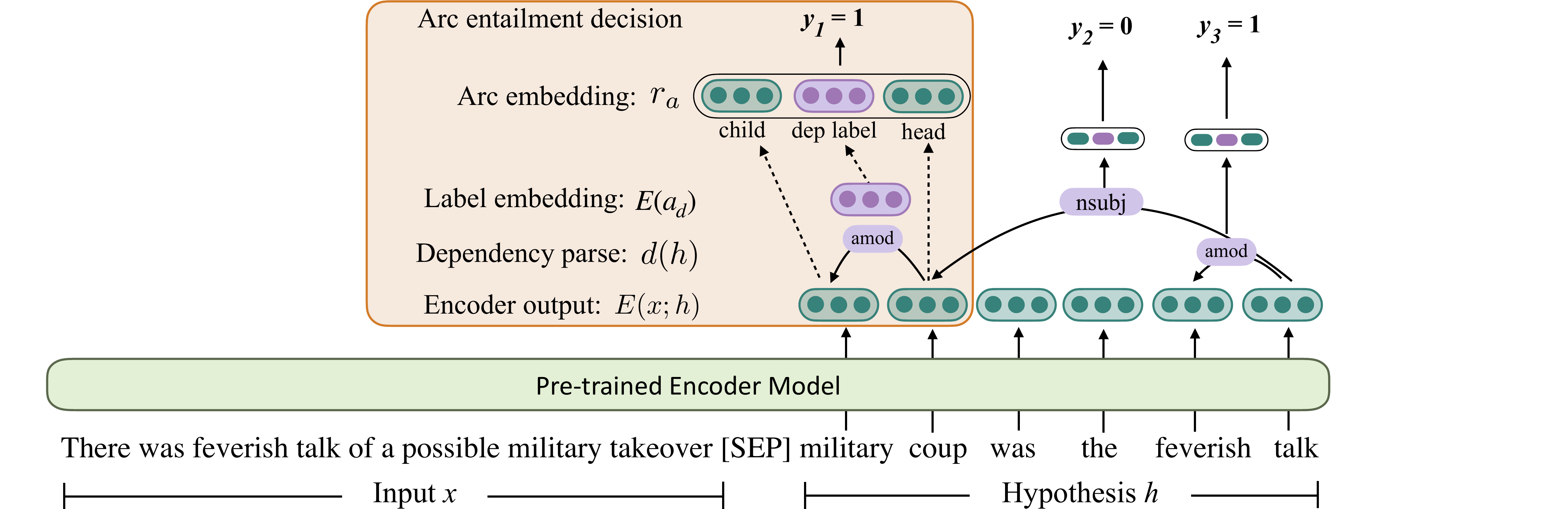}
    \caption{Overview of our dependency arc entailment model. The input (premise) sentence and output (or prefix of the output) are encoded with a pre-trained model. The embeddings of the head and tail of an arc are selected, concatenated with an encoding of the dependency label, and fed to a classification layer to render the judgment.} 
    \label{fig:model}
\end{figure*}


In some ways, our view of entailment is equivalent with the entailment of NLI settings \cite{bowman-etal-2015-snli,N18-1101}: if a hypothesis is entailed under the NLI definition, then all dependency arcs of the hypothesis must be entailed by our DAE definition. 
However, in our formulation, arc entailment is a 2-class classification task with labels $\in$ \emph{ \{entailed, non-entailed\}}. This means that arcs that would be \emph{neutral} or \emph{contradiction} in the generic entailment formulation are considered \emph{non-entailed} in our scenario.

\paragraph{Annotating arc entailment} To model this formulation, we require entailment annotations at the dependency arc level. However, there are several challenges associated with human annotation of arc entailment data. (1) Entailment is not truly a binary decision and is inherently subjective \cite{pavlick2019inherent}. (2) Entailment of an arc may be fundamentally unknowable or undefined if, for example, too much of the context has changed for such a judgment to make sense. (3) Annotators would need to understand the meaning of dependency labels and to be able to isolate the semantics of these individual arcs in sentences.

\begin{table}
\small
\centering
\begin{tabular}{p{0.95\linewidth}} \toprule
    P: Australian Prime Minister, John Howard, has made an unannounced visit to Iraq, according to the office of Iraqi transitional Prime Minister Ibrahim al-Jaafari.  \\
    H: Howard is a political representative of Australia. \\
    \bottomrule
\end{tabular} 
\caption{Example of premise (P) and hypothesis (H) with label \emph{entailment} from the dev set of entailment dataset RTE-2 \cite{bar2006second}.}
\label{table:entail-ex} 
\end{table}

Therefore, in this work, we take another approach, which is to \emph{automatically} label data from existing sources and outputs of generation models, which lets us collect large-scale data in a variety of domains. Specifically, we use paraphrase data to construct our dataset. Note however, that there is a fundamental difference between paraphrase pairs, which should be entailed in both directions, and past NLI data, which is forward-entailed by definition. For instance, the premise and hypothesis in Table~\ref{table:entail-ex} would classically be judged as entailed because \emph{political representative} is a hypernym of \emph{prime minister}, but the hypothesis is not a paraphrase of (even part of) the premise. 

As a result, our automatically-derived dataset captures a more restricted notion of entailment, primarily consisting of entailment relations that are symmetric in nature: arcs in the target sentence entailed by a source sentence also entail some part of the source.  However, this is actually closer to what is acceptable for generation models to produce in tasks such as summarization, and the dataset collected in such a manner is useful for downstream tasks, as we show in Section~\ref{sec:filtering-bad-gen}. Moreover, because our training and evaluation data will typically come from closely related sentences, we can sidestep the cases where judgments in our formalism become most difficult to define.


\section{Model} 
\label{sec:model}
Let $x$ be the input context, $h$ be a hypothesis produced by a generation model $\mathcal{G}$, and $d(h)$ be the set of arcs in the dependency parse of $h$. We want to predict the entailment decision for each arc $a\in d(h)$ with respect to the input $x$, i.e. $\mathcal{F}_a(a,x)$.

The overall model architecture of this dependency arc entailment model $\mathcal{F}_a$ is outlined in Figure~\ref{fig:model}.
First, we concatenate the input and the hypothesis. We use a pre-trained encoder model $E$ to obtain contextual representations for each token in the concatenated sequence. From these token level representations, we derive a representation for each dependency arc $a \in d(h)$: 
\begin{equation*}
    \mathbf{r}_a = [E(x;h)_{a_h}; E(x;h)_{a_c}; E(a_d)] 
\end{equation*} 
as shown in the inset in the figure. Here, $a_h$, $a_c$ are the token indices corresponding to the head word and the child word of dependency arc $a$, and $a_d$ is their corresponding dependency label, which is also embedded with $E$ (non-contextually).

Next, these arc representations are passed through a linear layer, followed by a softmax layer to obtain entailment label probabilities corresponding to each arc: $p(y \mid a;x) = \text{softmax}(W\mathbf{r}_a))$. 

This DAE network is trained using standard binary cross entropy loss and requires supervision on the arc entailment labels $y^* \in$ \emph{\{entailed, non-entailed\}} for the dependency arcs. Examples do not need to be fully labeled; training can use partial sets of annotations of arcs in $d(h)$. However, while using the DAE model in downstream tasks such as hypotheses reranking, entailment decisions for all arcs in the candidate hypothesis are required. 

\paragraph{Sentence-level factuality from dependency-level judgments}
We want to evaluate the factual consistency of each hypothesis $h$ with respect to input $x$, i.e. $\mathcal{F}(h, x)$. This is computed by combining arc-level entailment scores over the dependency arcs set $d(h)$ of the generated hypothesis:
\begin{equation*}
    \mathcal{F}(h, x) = \frac{1}{|d(h)|}\sum_{a \in d(h)}{\mathcal{F}_a(a, x)} 
\end{equation*}
We use the sentence-level score $\mathcal{F}(h, x)$ to rerank the generated hypotheses in Section~\ref{sec:filtering-bad-gen}.\footnote{According to DAE definition, an output is non-factual if  any of its arcs is non-entailed. However, min-pooling was very unstable, so we instead use mean-pooling in our experiments.}

\section{Automatic Dataset Creation}
\label{sec:data} 
We now describe our method for automatically collecting dependency-level DAE annotations from paraphrase or entailment corpora, avoiding manual annotation. In this creation process, we want data featuring a range of paraphrasing phenomenon, such as passivization, clausal reordering, synonym replacement, and more. Furthermore, we want a natural distribution of errors produced by generation models, such as wrong subject or objects for a verb or hallucination of new content.

We represent a single example in our dataset as a tuple $\big(x, h, \big\{(a_i, y_i^*)\big\}_{a_i \in d(h)}\big)$ . Here, $x$ is the input, $h$ is the hypothesis, $a_i$ denotes a single dependency arc in the hypothesis, and $y_i$ refers to the gold entailment label for that arc. To construct data of this form, we assume access to a paraphrase dataset $D$, containing pairs $(x,h^*)$ of input sentences and their corresponding gold paraphrases.\footnote{The paraphrase corpora we use in this work may come from automatic methods like backtranslation; however, we still assume that these are reliable gold-standard paraphrases.} Additionally, we employ a paraphrase generation model $\mathcal{G}_p$, which can output $k$ candidate paraphrases $\{h_1, h_2, ... h_k\}$ given an input $x$. These noisy paraphrases will be used to derive realistic examples of generation errors to contrast with gold paraphrases, using the following techniques.

\paragraph{Positive labels from gold paraphrases} Given a ground truth paraphrase, \textbf{we assume that every arc in the target side of the paraphrase $h^*$ is entailed by the source side $x$}. This is in line with our definition of arc entailment in Section~\ref{sec:DAE} and allows us to propagate sentence-level paraphrase judgements to arc-level entailment judgements. Because paraphrase datasets feature diverse linguistic phenomena, this approach leads to a range of positive examples. However, as described in Section~\ref{sec:DAE}, it is less likely to include arcs which are forward-entailed only (e.g., Table~\ref{table:entail-ex}).


\paragraph{Auto-derived labels from model generations} To find negative examples for entailment, we leverage the output generations $\{h_1, h_2, ... h_k\}$ of an automatic paraphrase model $\mathcal{G}_p$. These generations will include unseen arcs, which may be positively or negatively entailed.


Our key assumption here is that the outputs at the top of the beam are more likely to be factually correct, whereas outputs at the bottom of the beam are of lower quality and more prone to having factual inconsistencies. \textbf{We assume that \emph{new} arcs introduced in \emph{bad} model generations (i.e., bottom-most generations of the beam) are not entailed by the input.} 

We can then noisily label the generated paraphrases with a mix of positive (\emph{entailed}) and negative (\emph{non-entailed}) labels. We first construct a set of \emph{entailed} dependency arcs: this is a set containing all dependency arcs of the input and the gold paraphrase, i.e., $d(x) \cup d(h^*)$. Next, we annotate the dependency arcs of the bottom-most generations of the beam, say $\{h_{k}, h_{k-1}, h_{k-2}\}$, in the following way: 
\begin{equation*}
    y_i=
    \begin{cases}
      1 & \text{if}\ a_i \in d(x) \cup d(h^*) \\
      \text{not labeled} & \text{if}\ a_i \in d(h_1)\backslash  d(x) \cup d(h^*) \\
      0 & \text{otherwise} 
    \end{cases} 
\end{equation*}
The middle case here leaves arcs that are in $h_1$ but not $x$ or $h^*$ as unannotated. Such arcs are possibly factual under the model, coming from a high-quality generated output, but we do not have enough confidence to assign them a label. During model training, such unannotated arcs are ignored.

Finally, we also include the positive arcs from the 1-best hypothesis in our DAE data: $(x, h_1, \{a_i, 1 \})$ for arcs $a_i \in d(x) \cup d(h^*)$. This provides another source of hypothesis sentences with a slightly different distribution during model training.

\section{Intrinsic Evaluation of DAE}
Our experimental evaluation focuses on the following questions: (1) Does the automatic data collection result in a high quality training dataset? (2) Is the DAE model we train a good classifier? (3) Does DAE allow us to filter model generations and improve reliability of a generation system?

We construct our DAE training dataset using the methodology defined in Section~\ref{sec:data}. For this, we leverage the paraphrase pair dataset \textsc{Paranmt-50m} \cite{wieting-gimpel-2018-paranmt} as the base dataset $D$. We use the transformer-based encoder-decoder model for paraphrase generation from \citet{Goyal-Durrett:2020:Syntaxparaphrase} as $\mathcal{G}_p$. 
We use the paraphrase model to generate $10$ outputs for 20k sentence pairs from $D$. We use the Stanford CoreNLP library \cite{manning-EtAl:2014:P14-5} to extract enhanced dependencies from the outputs sentences. Then, using the strategy outlined in Section~\ref{sec:data}, we generate 100k training samples (sentence pairs), 3k dev samples and 3k test samples. From this dataset, we derive 520k training, 14k dev, and 22k dependency level annotations, which we evaluate on in Section~\ref{sec:intrinsic}. The entailed to not-entailed ratio is roughly 70-30 in this dataset.

\subsection{Dataset Quality} 
\begin{figure}[h]
\centering
    \includegraphics[trim=15mm 160mm 173mm 20mm,scale=0.44,clip]{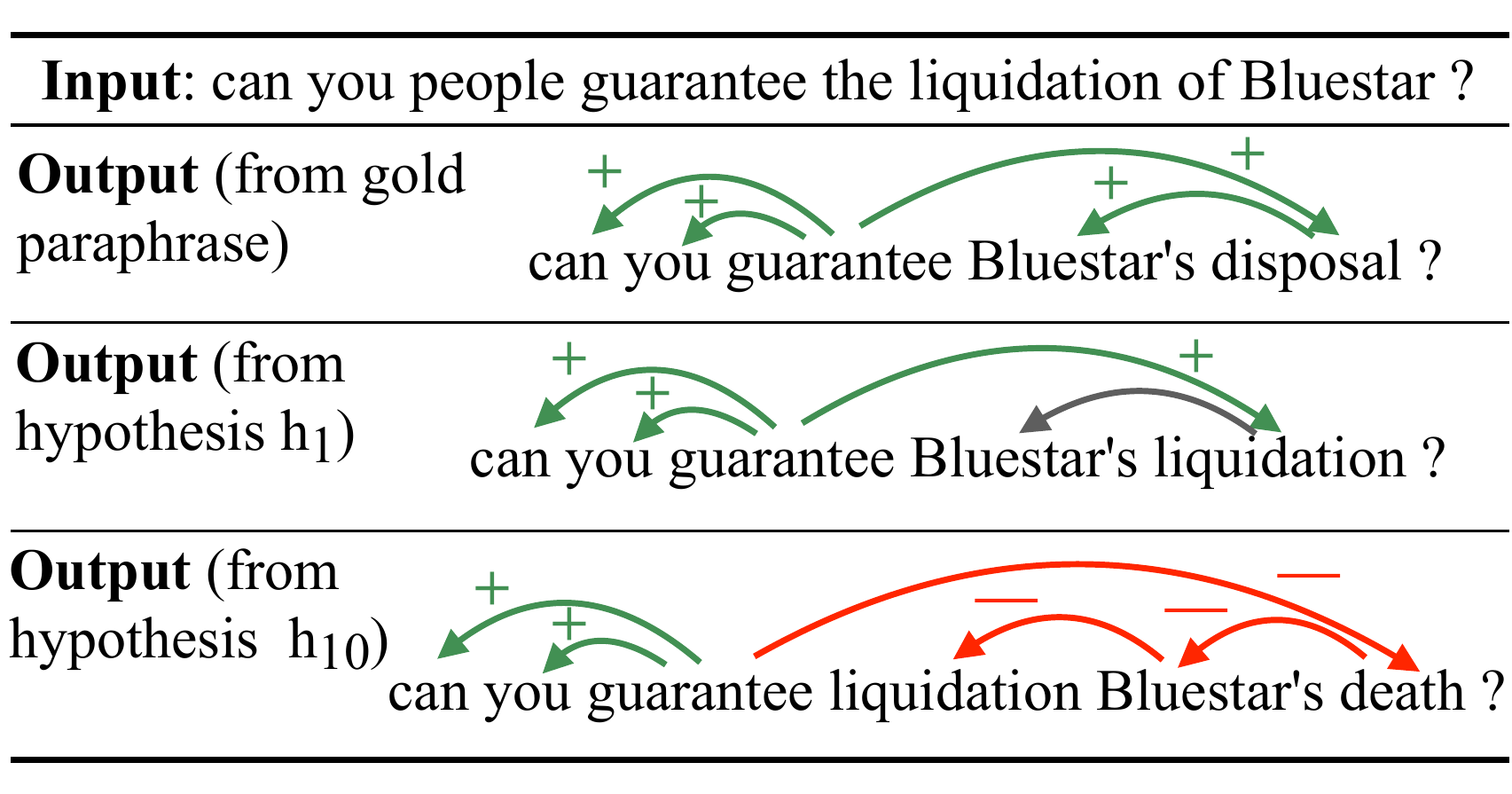} 
    \caption{Arc annotations from the automatic labelling strategy of Section \ref{sec:data}. Green (+) arcs are labelled \emph{entailed}, red (-) arcs are \emph{non-entailed}, and the gray arcs are unannotated.}
    \label{fig:dae-dev-ex}
\end{figure}

Before evaluating our modeling approach, we first evaluate whether the arc annotations in the training data follow the theoretical definition of DAE, outlined in Section \ref{sec:DAE}. Figure \ref{fig:dae-dev-ex} showcases examples from the dev set, corresponding to the same input example. We show positive \emph{entailed} arcs (in green), negative \emph{non-entailed} arcs (in red), and one unlabeled arc (in gray). Here, we can see that the gold paraphrase is important as it provides examples of valid synonym replacements, as well as other rephrasing of the input sentence. For negative examples, the examples from the bottom of the beam do indeed contain bad output and \emph{non-entailed} arcs.

\paragraph{Agreement with human labels} Next, we want to evaluate the quality of the auto-derived dataset by measuring its agreement with human annotations. For this, we manually annotated the dependency arc entailment labels for 100 sentence pairs from the dev set (consisting of 20 gold paraphrases and 80 generated paraphrases), according to our theoretical definition.  We compared these manual annotations (gold) with the auto-derived annotations for this set, and observed that the two annotations agreed 82\% of the time. This indicates that the automatic annotation strategy from Section \ref{sec:data} results in a high quality dataset. Further investigation into the disagreements between the manual and automatic labels revealed that false negatives included paraphrasing phenomena like synonym replacement, anaphora resolution during reordering, etc. We describe how to produce additional data to handle some of these cases later. On the other hand, false positives mainly consisted of exact arc matches in incorrect contexts. 


\subsection{Intrinsic Evaluation: DAE Classification} 
\label{sec:intrinsic}
\begin{table}[h]
\small
\centering
\begin{tabular}{r|cc}
\toprule
Model & Accuracy & F1 \\ \midrule
Majority label  & 72.7 & 84.2 \\
Lexical-match  & 74.2 & 78.1 \\
\textsc{Bert}  & 86.9 & 91.0 \\
\textsc{Electra}  & \textbf{88.4} & \textbf{92.1}\\
\bottomrule
\end{tabular}
\caption{Dependency-level performance of the different models on our held-out DAE examples constructed from paraphrase data. Results show that transformer based models outperform the baseline models.}
\label{table:intrinsic} 
\end{table}

\noindent
Next, we intrinsically evaluate the performance of the dependency arc entailment model, outlined in Section~\ref{sec:model}, on held-out data from our automatic labeling method. 
We test the performance of two pre-trained models: \textsc{Bert} (bert-base-uncased, 110M parameters) \cite{devlin2019bert} and \textsc{Electra} (electra-base-discriminator, 110M parameters) \cite{Clark2020ELECTRA}. We compare these models against a \textbf{majority label} baseline (\emph{entailment}) and an \textbf{lexical-match} baseline that predicts $y=\text{\emph{entailment}}$ if the arc (head, child and label)   in the output constitute a dependency arc in the input as well, and \emph{non-entailed} otherwise.

The performance of the different models is outlined in Table~\ref{table:intrinsic}. Our pre-trained transformer models perform substantially better than the baselines, with \textsc{Bert} achieving $86.9\%$ accuracy, and \textsc{Electra} with $88.4\%$. These models also outperform the lexical-match baseline, showing that the DAE models learn to do more than simple dependency arc matching. Henceforth, we use the best performing \textsc{Electra} model in all our experiments.

\subsection{Other Data Generation Methods} 
Besides the data generation procedure we proposed, there are other ways to synthetically generate noisy annotations for premise-hypothesis pairs \cite{zhang2019paws}. We investigate these from two perspectives: first, does our data generation process cover these phenomena well, and second, can these additional sources of data prove useful?

First, we explore a \textbf{word-swapping} technique similar to \citet{zhang2019paws}. Given a premise $x$, we form a hypothesis $x'$ by randomly swapping tokens that share a common part-of-speech tag to introduce errors. The intersection of the arcs in $d(x)$ and the modified sentence $d(x')$ are annotated as positive arcs ($y=\text{\emph{entailment}}$), whereas the newly created or changed arcs are annotated as negative ($y=\text{\emph{non-entailed}}$).

Our \textbf{synonym data} is noisily constructed in the same manner as the gold paraphrases, but targets synonym replacements specifically. We extract pairs $(x, h^*)$ from $D$ that generally maintain similar sentence structure between the two sentences,\footnote{We follow prior work \cite{Goyal-Durrett:2020:Syntaxparaphrase} and calculate structure similarity by aligning words in the input and target using GloVe \cite{pennington2014glove} and computing the average displacement of each word.} but with small lexical changes like synonym replacement. We assign a positive \emph{entailment} label to all arcs: $\big(x, h^*, \{(a, 1)\ \; \forall a \in d(h^*)\}\big)$.

To construct data with \textbf{hallucinations}, we modify an input sentence $x$, which we take as the \emph{hypothesis} by removing a randomly sampled span of contiguous tokens to derive a premise sentence $x'$. Then, the following DAE model annotations are derived: $\big(x', x, \{(a_i, 0)\; \forall \; a_i \in d(x) \backslash d(x')\}\big)$. Additionally, for each input sentence $x$, we extract another sentence $x'$ with the highest 1-gram overlap in the dataset. From this we derive, $\big(x, x', \{(a_i, 0)\; \forall \; a_i \in d(x')\}\big)$.

\begin{table}[h]
\small
\centering
\begin{tabular}{r|cccc}
\toprule
& \multicolumn{4}{c}{Test set}\\
Model Training Source & WS & AD & S & H \\ \midrule
Word-swapping (WS) & 98.5 & 71.6 & 29.6 & 80.0 \\
Auto-derived (AD) & 90.2 & 88.4 & 82.9 & 74.8 \\ 
 + synonyms (S) & 90.5 & 88.0 & 96.0 & 73.9 \\
 + hallucinations (H) & 92.4 & 87.8 & 96.9 & 97.6 \\
\bottomrule
\end{tabular}
\caption{Comparison of different training data methodologies. Our method with augmentations (AD+S+H) performs well on all categories.} 
\label{table:data-ablations} 
\end{table}

Table \ref{table:data-ablations} shows a comparison of word-swapping with our method (AD), and variants of our method augmented with synonyms and hallucinations. Note that the model trained on word swapped data performs well on a similarly constructed held-out set, but not on the test data for synonym data and auto-derived data. This indicates that artificially constructed data with rule based error introduction does not cover the space of generation possibilities. 
On the other hand, the model trained on our auto-derived dataset performs well across both artificial and actual generation data, thereby covering a larger range of entailment possibilities. Additional augmentation of synonym- and hallucination-specific data improves the performance further on the respective test sets while retaining the performance on generic entailment data. Henceforth, we use the (AD + S) model for our experiments. 

\section{Extrinsic Evaluation: Filtering Bad Generations} 
\label{sec:filtering-bad-gen}

Moving beyond the dependency-level inference task, we now want to evaluate the \emph{sentence-level} performance of our model formulation. Namely, can it usefully reject erroneous generations produced by models for summarization (Section~\ref{sec:summarization-eval}) and paraphrasing (Section~\ref{sec:paraphrasing-eval})?

\subsection{Summary Ranking} 
\label{sec:summarization-eval}

We perform our evaluation on an abstractive summarization test dataset introduced in \citet{falke2019ranking} and used in other previous work. It contains $373$ test samples, each containing an input source sentence from CNN/DM and two summary sentences covering the same content generated using the model from \citet{chen2018fast}. One of these summary sentences is factually correct and the other is factually incorrect. The evaluation protocol measures how often the correct summary is scored higher than the incorrect summary for each candidate scoring technique. 
We compare against the following baselines:
\begin{enumerate}[leftmargin=*]
    \item \textbf{NLI models}: Following \citet{falke2019ranking}, we use entailment predictions of NLI models to rerank candidates. We compare the performance of pretrained encoders (\textsc{Bert}, \textsc{RoBerta} and \textsc{Electra}) fine-tuned on the MNLI dataset \cite{N18-1101}.\footnote{We fine-tune the \textsc{Bert} and \textsc{Electra} models ourselves (details in the Appendix), improving on the results from \citet{falke2019ranking}. We use the fine-tuned \textsc{RoBERTa} model released by AllenNLP (\url{https://allennlp.org/}).} 
    \item \textbf{Question Generation and Answering}: \citet{wang2020asking} propose an automatic evaluation metric QAGS that scores each summary by first generating questions pertaining to the summary content, and then comparing the answers to those questions in both the source sentence and the generated summary.
    \item \textbf{Rule-based}: We score each summary sentence as the fraction of dependency arcs in the output that are common with the input sentence. In case both the correct and the incorrect sentence get the same score, we break ties randomly.
\end{enumerate}

\begin{table}[t]
\small
\centering
\begin{tabular}{r|c}
\toprule
Model & Reranking Acc. \\ \midrule
\textsc{BERT} (MNLI)  & 72.3 \\
\textsc{RoBERTa} (MNLI) & 78.3  \\
\textsc{Electra} (MNLI) & 82.0 \\ \midrule
QAGS & 72.1 \\ \midrule
Rule-based dependency & 74.8   \\ 
DAE (ours) & 83.6 \\ \midrule
Human & 83.9 \\
\bottomrule
\end{tabular}
\caption{Performance of the different models at the summary reranking task. The human baseline is reported in \citet{falke2019ranking}. The proposed DAE model performs on par or better than prior works and comes close to human performance.} 
\label{table:summ-reranking} 
\end{table}

Table \ref{table:summ-reranking} outlines the performance of the different models. The results show that the dependency arc entailment model outperforms the sentence-level NLI models and also the question generation and answering formulation (QAGS). Furthermore, the performance of our DAE model is close to the human performance reported in \citet{falke2019ranking}. 
Interestingly, the rule-based dependency model also outperforms certain NLI models and QAGS, indicating that these more complex models may fail to capture straightforward lexical relations.

\begin{figure}[t!]
\centering
    \includegraphics[trim=48mm 153mm 120mm 50mm,scale=0.48,clip]{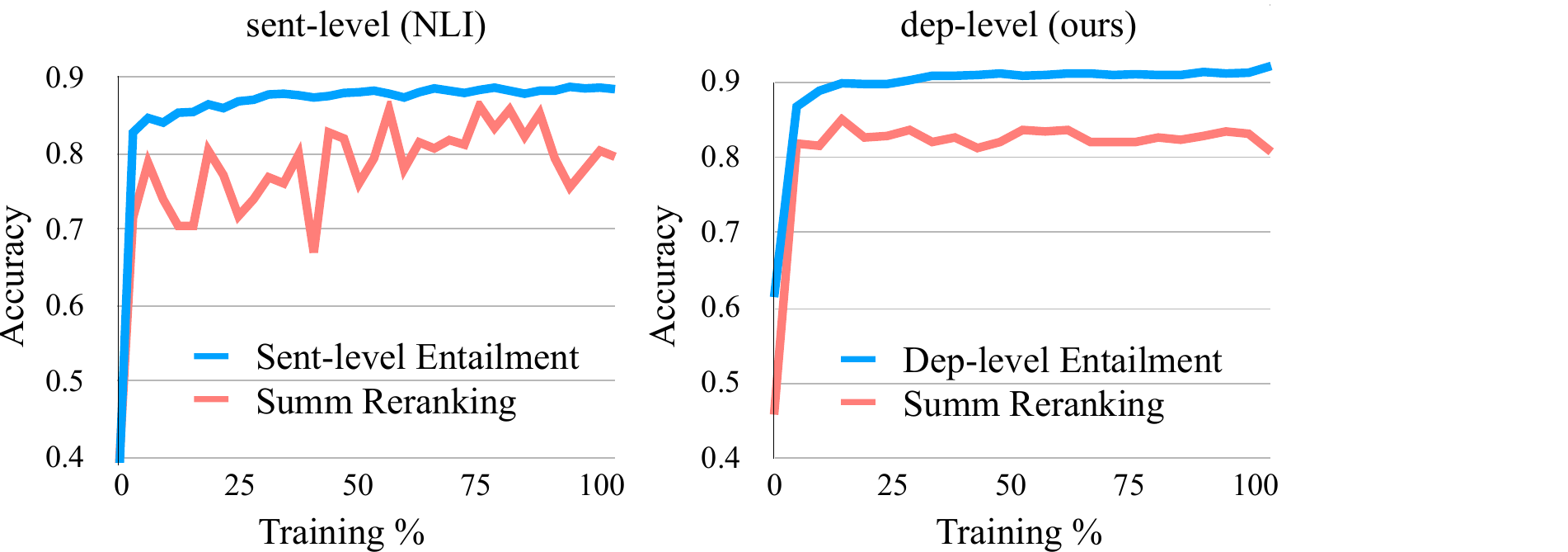}
    \caption{Performance of the \textsc{Electra}-based MNLI model and the DAE model. The figure shows a much higher variance in reranking accuracy for the MNLI model, suggesting that the task-specific performance is not correlated with reranking performance.}
    \label{fig:variance-mnli-DAE} 
\end{figure}

During our experimentation, we observed large variance in the performance of the NLI models at the reranking task with respect to their performance at the intrinsic entailment task. To illustrate this, in Figure~\ref{fig:variance-mnli-DAE}, we plot the summarization reranking performance of the two model against the intrinsic task performance at different stages of the training. For DAE, the intrinsic task performance is reported by the dependency-level entailment classification accuracy, and for the MNLI model, we report the classification accuracy on the sentence-level MNLI entailment task. The graph shows a high variance in the summary reranking performance, with stable increase in the MNLI intrinsic task performance at different time steps.\footnote{Note that the best performance of the MNLI model on summary reranking is better than the best performance of the DAE model; however, it did not coincide with the task-level best performance for our particular hyperparameter choice.} This indicates that the general entailment task solves a fundamentally different problem than factuality.
By contrast, the DAE model performance on the summarization reranking task is more stable.

\subsection{Paraphrase Ranking} 
\label{sec:paraphrasing-eval}
Next, we evaluate the DAE model in the paraphrasing setting. To do this, first, we create a test set, similar to the summarization test set from \citet{falke2019ranking}. We use the transformer based seq2seq model \cite{Goyal-Durrett:2020:Syntaxparaphrase} to obtain $10$ candidate paraphrases for $100$ input sentences from the \textsc{ParaNMT-50m} dataset. We manually assign a label $y \in $ $\{$\emph{factual}, \emph{not factual}$\}$ to each input, candidate pair. Then for each input sentence, we randomly selected one correct and one incorrect paraphrase. This sentence triplet is used for our reranking experiments. 

\begin{table}[h]
\small
\centering
\begin{tabular}{c|cc}
\toprule
Model & Reranking Acc \\ \midrule
MNLI (\textsc{Electra}) & 79.0 \\
DAE (\textsc{Electra})  & 79.0  \\
\bottomrule
\end{tabular}
\caption{Performance on the paraphrase reranking task. The DAE performs on par with the NLI based model.}
\label{table:para-rerank}
\end{table}
Table \ref{table:para-rerank} compares the performance of the MNLI-based  model and the DAE models. Here, both are \textsc{Electra} based models; these are shown to be the best performing models in the previous sections. The results show that in this setting, the MNLI model and the DAE model perform similarly. Closer inspection of this data revealed that our model is biased towards predicting the label \emph{entailment} for arcs that are common with the input, possibly because we are evaluating the same generation model that was used to produce our arc entailment data, and our model is therefore biased towards predicting \emph{non-entailed} for arcs that are not present in the input. This poses a somewhat adversarial setting for our DAE model.

\section{Analysis} 

\begin{figure*}[h]
\centering
    \includegraphics[trim=14mm 77mm 0mm 34mm,scale=0.46,clip]{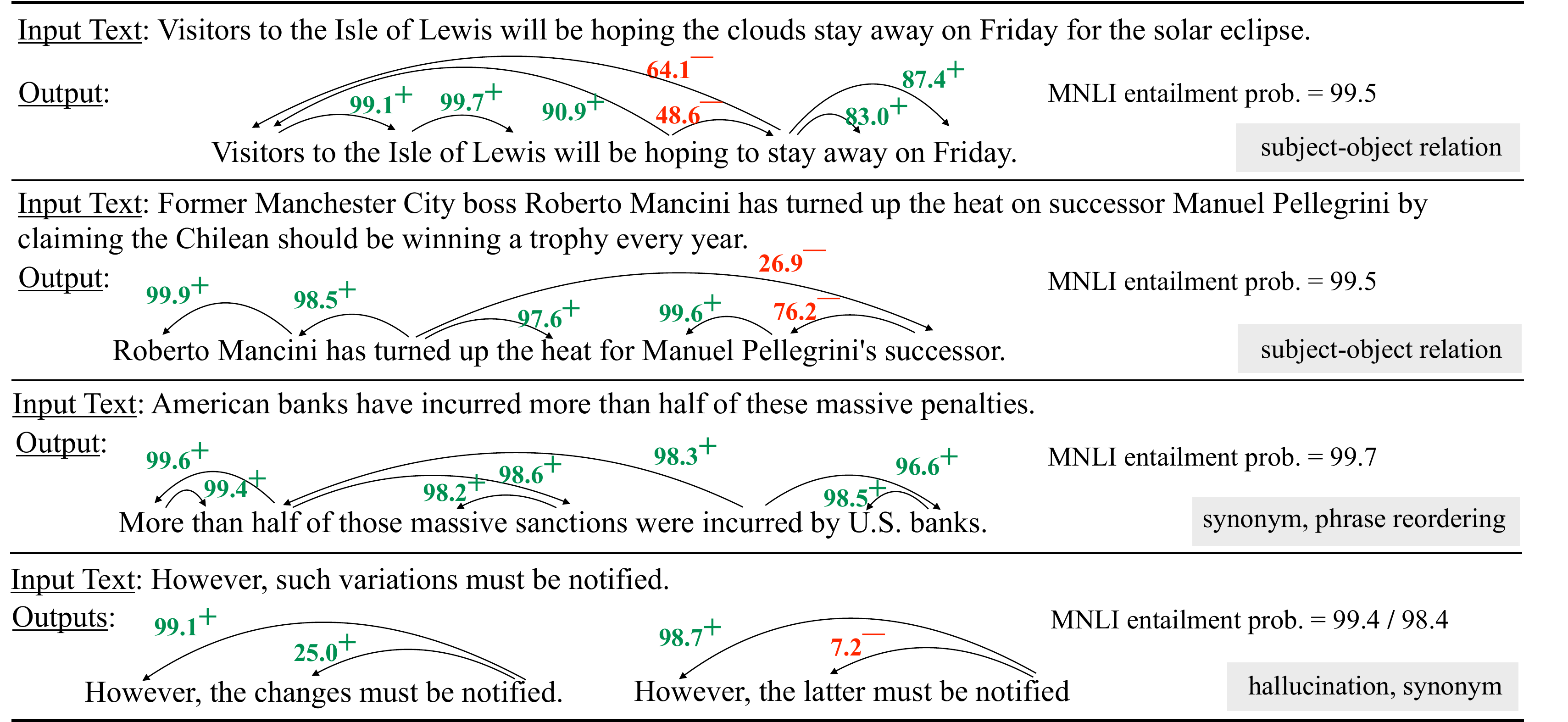}
    \caption{Individual arc entailment probabilities for arcs in output sentences from the summarization test set \cite{falke2019ranking} and the paraphrase test set. The $+/-$ superscript signifies the gold label for that arc. Our DAE model is able to localize errors in the output. Compared to this, the MNLI model computes a high entailment score for all arcs that are lexically similar.} 
    \label{fig:example-summ}
\end{figure*}
\subsection{Dependency- vs.~sentence-level modeling} 

Although our DAE model has shown strong performance, we have not yet performed a direct apples-to-apples comparison of DAE versus a sentence-level model \emph{when trained on the same sentences.}

\paragraph{For MNLI} We construct DAE data from the sentence-level entailment data as follows. First, we extract 10k examples from the MNLI data which have the label \emph{entailment}. This is considered as the source data $D'$. We use a paraphrase model (transformer seq2seq \cite{Goyal-Durrett:2020:Syntaxparaphrase}) and the technique outlined in Section~\ref{sec:data} to extract auto-derived labels from $D'$. This gives us 42k training examples for training the DAE model. We compare this against an MNLI model trained on the original sentence-level entailment task with the same number of examples (42k). 

\paragraph{For \textsc{ParaNmt}} For this dataset, we do not have negative ($y = \text{\emph{contradiction}}$) annotations at the sentence-level. We derive these from our DAE training data as follows: we consider all pairs of sentences in the original dataset ($x,h^*$) as positive sentences ($y=1$), in addition to any pair of the form ($x,x$). We treat the three generated sentences at the bottom of the beam as negative sentences, meaning that the model is trained to distinguish gold paraphrases from model generations.

\begin{table}[h]
\small
\centering
\begin{tabular}{c|cc|cc}
\toprule
Model & \multicolumn{2}{c}{ParaNMT} & \multicolumn{2}{|c}{MNLI}\\ \midrule
& Summ & Para & Summ & Para \\ \midrule
sent-level  &  73.9 & 58.0 &  68.8 & 64.0\\ 
dep-level & 83.6 & 79.0  & 78.5 & 79.0\\
\bottomrule
\end{tabular}
\caption{Comparison of the sentence-level and dependency-level formulations. On similarly sized training datasets, the dependency-level formulation outperforms the sentence-level formulation for both types of data sources considered.} 
\label{table:sent-vs-dep} 
\end{table}

Table~\ref{table:sent-vs-dep} outlines these results. For the paraphrase dataset, we see that the artificially constructed sentence-level dataset does not yield a good sentence-level discriminator. However, our dependency-level annotations \emph{can} form an effective training set. The results on MNLI show that our dependency-level formulation performs better than sentence-level when trained on the same amount of data and is therefore more closely related to the factuality task than past entailment formulations.

\subsection{Qualitative Evaluation} 
\paragraph{Error Localization}
Since the DAE formulation computes individual entailment scores for all arcs in the dependency tree structure, it is possible to localize errors in the generated summary or paraphrase. We present examples of input sentences, generated text, and arc entailment scores for a few examples in Figure~\ref{fig:example-summ}. For each input and output pair, we show the individual scores for the dependency arcs in the output sentence. Additionally, we report the MNLI score for the same example.

The illustrative examples show that the DAE model is capable of localizing errors where erroneous subject-object pairs were constructed, even when these are the only errors. These errors are tougher for the MNLI model to catch, which evaluates the whole sentence and is prone to lexical overlap biases \cite{zhou2020towards}.  In the third example, from our paraphrase setting, we see that the model is able to recognize synonym replacement as a valid re-writing. However, for the last example, the model cannot perform this same judgement for the \emph{variations $\rightarrow$ changes} replacement. Although, note that the model scores it higher than a erroneous replacement of the same word (\emph{variations $\rightarrow$ latter}). This shows that the DAE model is able to rank sentences that incorporate the similar type of re-writing/editing.
However, we observed that the model has different error rates for different types of re-writing changes. For example, it is better at identifying text hallucination, or cases where the subject object relation between words change, but has comparatively lesser accuracy over changes such as synonym replacements. Therefore, it may not be ideal for settings where different re-writing types need to be compared. 

\paragraph{Limitations}
\label{sec:limitations}

We comment on a few limitations of our approach. First, arcs in our dependency-based formalism are not marked with negation or quantification; these must be handled via the contextualization of the hypothesis sentence rather than in the semantic representation. Second, our method cannot identify arcs that are missing from the input. For instance, consider the following premise: \emph{ In the morning, he goes jogging} and hypothesis: \emph{In the morning}. 
Here, the hypothesis does not contain critical information from the source sentence; however, since all the \emph{present} arcs are entailed by the source, our model would  give this a high score. 


Furthermore, our model is trained on the \textsc{ParaNMT-50m} dataset, which itself is constructed through a noisy backtranslation process. Therefore, we rely on noisy \emph{gold} data for constructing our model. We believe that better quality paraphrase pairs would lead to a better quality model.

\section{Related Work} 
Recent work in addressing faithfulness of text generations can be broadly divided into three groups: structured information based, multi-task formulations, and post-processing methods. The first group leverages structured knowledge, like Open IE triples \cite{cao2018faithful, goodrich2019assessing}, dependency trees  \cite{song2018structure}, or generated semantic roles \cite{fan18irasl} as additional input for generation. However, incorporation of these as additional embeddings in model architectures does not explain \emph{how} these influence model generations. The second group leverages multi-task formulations and trains the generation model jointly with other factuality-related tasks, such as NLI entailment and question generation \cite{guo2018soft}. Other work additionally incorporates a reward for generating summaries entailed by the the input \cite{li2018ensure,pasunuru2018multi}. Our approach can be used to rank/filter outputs from any generation model in a black-box way, without additional augmentation or retraining. 

In post-processing approaches, recent work has explored NLI-based \cite{falke2019ranking, maynez2020faithfulness} post-generation filtering or ranking of output summaries. Our dependency-level models perform on par with these approaches, while additionally localizing the error in the generations. Other work \cite{durmus2020feqa, wang2020asking} has looked at using question generation and answering to reveal factual inconsistencies in generated text. However, more work is needed to figure out how to make these approaches reliable and broad coverage, as they primarily focus on specific factors like noun phrases and named entities.

\section{Conclusion} 
In this work, we propose the dependency arc entailment formulation to identify factual errors in generated text in a more fine-grained manner. We show that the proposed formulation outperforms past approaches, while additionally providing an interpretable error analysis. 
\section*{Acknowledgments}

Thanks to Katrin Erk for providing feedback on a draft of this work, as well to the anonymous reviewers for their helpful comments. This work was partially supported by NSF Grant IIS-1814522, a gift from Salesforce Inc, and an equipment grant from NVIDIA. The authors acknowledge the Texas Advanced Computing Center (TACC) at The University of Texas at Austin for providing HPC resources used to conduct this research.

\bibliography{anthology,emnlp2020}

\begin{thebibliography}{40}
\expandafter\ifx\csname natexlab\endcsname\relax\def\natexlab#1{#1}\fi

\bibitem[{Banarescu et~al.(2013)Banarescu, Bonial, Cai, Georgescu, Griffitt,
  Hermjakob, Knight, Koehn, Palmer, and Schneider}]{banarescu2013abstract}
Laura Banarescu, Claire Bonial, Shu Cai, Madalina Georgescu, Kira Griffitt, Ulf
  Hermjakob, Kevin Knight, Philipp Koehn, Martha Palmer, and Nathan Schneider.
  2013.
\newblock {Abstract meaning representation for sembanking}.
\newblock In \emph{Proceedings of the 7th linguistic annotation workshop and
  interoperability with discourse}, pages 178--186.

\bibitem[{Bar-Heim et~al.(2006)Bar-Heim, Dagan, Dolan, Ferro, Giampiccolo,
  Magnini, and Szpektor}]{bar2006second}
Roy Bar-Heim, Ido Dagan, Bill Dolan, Lisa Ferro, Danilo Giampiccolo, Bernardo
  Magnini, and Idan Szpektor. 2006.
\newblock {The Second PASCAL Recognizing Textual Entailment Challenge}.
\newblock In \emph{Proceedings of the Second PASCAL Challenges Workshop on
  Recognising Textual Entailment, Venice, Italy}, page~49.

\bibitem[{Bowman et~al.(2015)Bowman, Angeli, Potts, and
  Manning}]{bowman-etal-2015-snli}
Samuel~R. Bowman, Gabor Angeli, Christopher Potts, and Christopher~D. Manning.
  2015.
\newblock {A large annotated corpus for learning natural language inference}.
\newblock In \emph{Empirical Methods in Natural Language Processing (EMNLP)}.

\bibitem[{Cao et~al.(2018)Cao, Wei, Li, and Li}]{cao2018faithful}
Ziqiang Cao, Furu Wei, Wenjie Li, and Sujian Li. 2018.
\newblock {Faithful to the original: Fact aware neural abstractive
  summarization}.
\newblock In \emph{Thirty-Second AAAI Conference on Artificial Intelligence}.

\bibitem[{Chen and Bansal(2018)}]{chen2018fast}
Yen-Chun Chen and Mohit Bansal. 2018.
\newblock {Fast Abstractive Summarization with Reinforce-Selected Sentence
  Rewriting}.
\newblock In \emph{Proceedings of the 56th Annual Meeting of the Association
  for Computational Linguistics (Volume 1: Long Papers)}.

\bibitem[{Clark et~al.(2020)Clark, Luong, Le, and Manning}]{Clark2020ELECTRA}
Kevin Clark, Minh-Thang Luong, Quoc~V. Le, and Christopher~D. Manning. 2020.
\newblock \href {https://openreview.net/forum?id=r1xMH1BtvB} {{ELECTRA:
  Pre-training Text Encoders as Discriminators Rather Than Generators}}.
\newblock In \emph{International Conference on Learning Representations}.

\bibitem[{Devlin et~al.(2019)Devlin, Chang, Lee, and
  Toutanova}]{devlin2019bert}
Jacob Devlin, Ming-Wei Chang, Kenton Lee, and Kristina Toutanova. 2019.
\newblock {BERT: Pre-training of Deep Bidirectional Transformers for Language
  Understanding}.
\newblock In \emph{Proceedings of the 2019 Conference of the North American
  Chapter of the Association for Computational Linguistics: Human Language
  Technologies, Volume 1 (Long and Short Papers)}.

\bibitem[{Dong et~al.(2019)Dong, Yang, Wang, Wei, Liu, Wang, Gao, Zhou, and
  Hon}]{dong2019unified}
Li~Dong, Nan Yang, Wenhui Wang, Furu Wei, Xiaodong Liu, Yu~Wang, Jianfeng Gao,
  Ming Zhou, and Hsiao-Wuen Hon. 2019.
\newblock {Unified language model pre-training for natural language
  understanding and generation}.
\newblock In \emph{Advances in Neural Information Processing Systems}, pages
  13042--13054.

\bibitem[{Durmus et~al.(2020)Durmus, He, and Diab}]{durmus2020feqa}
Esin Durmus, He~He, and Mona Diab. 2020.
\newblock {FEQA: A Question Answering Evaluation Framework for Faithfulness
  Assessment in Abstractive Summarization}.
\newblock In \emph{Proceedings of the 58th Annual Meeting of the Association
  for Computational Linguistics}.

\bibitem[{Falke et~al.(2019)Falke, Ribeiro, Utama, Dagan, and
  Gurevych}]{falke2019ranking}
Tobias Falke, Leonardo~FR Ribeiro, Prasetya~Ajie Utama, Ido Dagan, and Iryna
  Gurevych. 2019.
\newblock {Ranking generated summaries by correctness: An interesting but
  challenging application for natural language inference}.
\newblock In \emph{Proceedings of the 57th Annual Meeting of the Association
  for Computational Linguistics}, pages 2214--2220.

\bibitem[{Fan et~al.(2018)Fan, Yu, and Wang}]{fan18irasl}
Lisa Fan, Dong Yu, and Lu~Wang. 2018.
\newblock {Robust Neural Abstractive Summarization Systems and Evaluation
  against Adversarial Information}.
\newblock In \emph{Workshop on Interpretability and Robustness in Audio,
  Speech, and Language (IRASL)}. Neural Information Processing Systems.

\bibitem[{FitzGerald et~al.(2018)FitzGerald, Michael, He, and
  Zettlemoyer}]{fitzgerald2018large}
Nicholas FitzGerald, Julian Michael, Luheng He, and Luke Zettlemoyer. 2018.
\newblock {Large-Scale QA-SRL Parsing}.
\newblock In \emph{Proceedings of the 56th Annual Meeting of the Association
  for Computational Linguistics (Volume 1: Long Papers)}, pages 2051--2060.

\bibitem[{Goodrich et~al.(2019)Goodrich, Rao, Liu, and
  Saleh}]{goodrich2019assessing}
Ben Goodrich, Vinay Rao, Peter~J Liu, and Mohammad Saleh. 2019.
\newblock {Assessing the factual accuracy of generated text}.
\newblock In \emph{Proceedings of the 25th ACM SIGKDD International Conference
  on Knowledge Discovery \& Data Mining}, pages 166--175.

\bibitem[{Goyal and Durrett(2020)}]{Goyal-Durrett:2020:Syntaxparaphrase}
Tanya Goyal and Greg Durrett. 2020.
\newblock {Neural Syntactic Preordering for Controlled Paraphrase Generation}.
\newblock In \emph{Proceedings of the 58th Annual Meeting of the Association
  for Computational Linguistics}.

\bibitem[{Guo et~al.(2018)Guo, Pasunuru, and Bansal}]{guo2018soft}
Han Guo, Ramakanth Pasunuru, and Mohit Bansal. 2018.
\newblock {Soft Layer-Specific Multi-Task Summarization with Entailment and
  Question Generation}.
\newblock In \emph{Proceedings of the 56th Annual Meeting of the Association
  for Computational Linguistics (Volume 1: Long Papers)}, pages 687--697.

\bibitem[{Gururangan et~al.(2018)Gururangan, Swayamdipta, Levy, Schwartz,
  Bowman, and Smith}]{gururangan2018annotation}
Suchin Gururangan, Swabha Swayamdipta, Omer Levy, Roy Schwartz, Samuel Bowman,
  and Noah~A Smith. 2018.
\newblock {Annotation Artifacts in Natural Language Inference Data}.
\newblock In \emph{Proceedings of the 2018 Conference of the North American
  Chapter of the Association for Computational Linguistics: Human Language
  Technologies, Volume 2 (Short Papers)}.

\bibitem[{Kry{ś}ci{ń}ski et~al.(2019)Kry{ś}ci{ń}ski, Keskar, McCann, Xiong,
  and Socher}]{kryscinski2019neural}
Wojciech Kry{ś}ci{ń}ski, Nitish~Shirish Keskar, Bryan McCann, Caiming Xiong,
  and Richard Socher. 2019.
\newblock {Neural text summarization: A critical evaluation}.
\newblock In \emph{Proceedings of the 2019 Conference on Empirical Methods in
  Natural Language Processing and the 9th International Joint Conference on
  Natural Language Processing (EMNLP-IJCNLP)}, pages 540--551.

\bibitem[{Lewis et~al.(2020)Lewis, Liu, Goyal, Ghazvininejad, Mohamed, Levy,
  Stoyanov, and Zettlemoyer}]{Lewis2019BARTDS}
Mike Lewis, Yinhan Liu, Naman Goyal, Marjan Ghazvininejad, Abdelrahman Mohamed,
  Omer Levy, Ves Stoyanov, and Luke Zettlemoyer. 2020.
\newblock {BART: Denoising Sequence-to-Sequence Pre-training for Natural
  Language Generation, Translation, and Comprehension}.
\newblock In \emph{Proceedings of the 58th Annual Meeting of the Association
  for Computational Linguistics}.

\bibitem[{Li et~al.(2018)Li, Zhu, Zhang, and Zong}]{li2018ensure}
Haoran Li, Junnan Zhu, Jiajun Zhang, and Chengqing Zong. 2018.
\newblock {Ensure the correctness of the summary: Incorporate entailment
  knowledge into abstractive sentence summarization}.
\newblock In \emph{Proceedings of the 27th International Conference on
  Computational Linguistics}, pages 1430--1441.

\bibitem[{Manning et~al.(2014)Manning, Surdeanu, Bauer, Finkel, Bethard, and
  McClosky}]{manning-EtAl:2014:P14-5}
Christopher~D. Manning, Mihai Surdeanu, John Bauer, Jenny Finkel, Steven~J.
  Bethard, and David McClosky. 2014.
\newblock \href {http://www.aclweb.org/anthology/P/P14/P14-5010} {The
  {Stanford} {CoreNLP} natural language processing toolkit}.
\newblock In \emph{Association for Computational Linguistics (ACL) System
  Demonstrations}, pages 55--60.

\bibitem[{Mao et~al.(2019)Mao, Majumder, McAuley, and
  Cottrell}]{mao2019improving}
Huanru~Henry Mao, Bodhisattwa~Prasad Majumder, Julian McAuley, and Garrison
  Cottrell. 2019.
\newblock {Improving Neural Story Generation by Targeted Common Sense
  Grounding}.
\newblock In \emph{Proceedings of the 2019 Conference on Empirical Methods in
  Natural Language Processing and the 9th International Joint Conference on
  Natural Language Processing (EMNLP-IJCNLP)}.

\bibitem[{Maynez et~al.(2020)Maynez, Narayan, Bohnet, and
  McDonald}]{maynez2020faithfulness}
Joshua Maynez, Shashi Narayan, Bernd Bohnet, and Ryan McDonald. 2020.
\newblock {On Faithfulness and Factuality in Abstractive Summarization}.
\newblock In \emph{Proceedings of the 58th Annual Meeting of the Association
  for Computational Linguistics}.

\bibitem[{Mel'\v{c}uk(1988)}]{Melcuk1988}
Igor Mel'\v{c}uk. 1988.
\newblock {Dependency Syntax: Theory and Practice}.
\newblock State University of New York Press.

\bibitem[{Michael et~al.(2018)Michael, Stanovsky, He, Dagan, and
  Zettlemoyer}]{michael2018crowdsourcing}
Julian Michael, Gabriel Stanovsky, Luheng He, Ido Dagan, and Luke Zettlemoyer.
  2018.
\newblock {Crowdsourcing Question-Answer Meaning Representations}.
\newblock In \emph{Proceedings of the 2018 Conference of the North American
  Chapter of the Association for Computational Linguistics: Human Language
  Technologies, Volume 2 (Short Papers)}, pages 560--568.

\bibitem[{Pasunuru and Bansal(2018)}]{pasunuru2018multi}
Ramakanth Pasunuru and Mohit Bansal. 2018.
\newblock {Multi-Reward Reinforced Summarization with Saliency and Entailment}.
\newblock In \emph{Proceedings of the 2018 Conference of the North American
  Chapter of the Association for Computational Linguistics: Human Language
  Technologies, Volume 2 (Short Papers)}.

\bibitem[{Pavlick and Kwiatkowski(2019)}]{pavlick2019inherent}
Ellie Pavlick and Tom Kwiatkowski. 2019.
\newblock {Inherent Disagreements in Human Textual Inferences}.
\newblock \emph{Transactions of the Association for Computational Linguistics}.

\bibitem[{Pennington et~al.(2014)Pennington, Socher, and
  Manning}]{pennington2014glove}
Jeffrey Pennington, Richard Socher, and Christopher~D. Manning. 2014.
\newblock \href {http://www.aclweb.org/anthology/D14-1162} {{GloVe: Global
  Vectors for Word Representation}}.
\newblock In \emph{Empirical Methods in Natural Language Processing (EMNLP)}.

\bibitem[{Radford et~al.(2019)Radford, Wu, Child, Luan, Amodei, and
  Sutskever}]{radford2019language}
Alec Radford, Jeff Wu, Rewon Child, David Luan, Dario Amodei, and Ilya
  Sutskever. 2019.
\newblock {Language Models are Unsupervised Multitask Learners}.

\bibitem[{Shen et~al.(2020)Shen, Chang, Su, Zhou, and Klakow}]{shen2020neural}
Xiaoyu Shen, Ernie Chang, Hui Su, Jie Zhou, and Dietrich Klakow. 2020.
\newblock {Neural Data-to-Text Generation via Jointly Learning the Segmentation
  and Correspondence}.
\newblock In \emph{Proceedings of the 58th Annual Meeting of the Association
  for Computational Linguistics}.

\bibitem[{Song et~al.(2018)Song, Zhao, and Liu}]{song2018structure}
Kaiqiang Song, Lin Zhao, and Fei Liu. 2018.
\newblock {Structure-Infused Copy Mechanisms for Abstractive Summarization}.
\newblock In \emph{Proceedings of the 27th International Conference on
  Computational Linguistics}, pages 1717--1729.

\bibitem[{Wang et~al.(2020)Wang, Cho, and Lewis}]{wang2020asking}
Alex Wang, Kyunghyun Cho, and Mike Lewis. 2020.
\newblock {Asking and Answering Questions to Evaluate the Factual Consistency
  of Summaries}.
\newblock In \emph{Proceedings of the 58th Annual Meeting of the Association
  for Computational Linguistics}.

\bibitem[{White et~al.(2016)White, Reisinger, Sakaguchi, Vieira, Zhang,
  Rudinger, Rawlins, and Van~Durme}]{white2016universal}
Aaron~Steven White, Drew Reisinger, Keisuke Sakaguchi, Tim Vieira, Sheng Zhang,
  Rachel Rudinger, Kyle Rawlins, and Benjamin Van~Durme. 2016.
\newblock {Universal decompositional semantics on universal dependencies}.
\newblock In \emph{Proceedings of the 2016 Conference on Empirical Methods in
  Natural Language Processing}, pages 1713--1723.

\bibitem[{Wieting and Gimpel(2018)}]{wieting-gimpel-2018-paranmt}
John Wieting and Kevin Gimpel. 2018.
\newblock \href {https://doi.org/10.18653/v1/P18-1042} {{P}ara{NMT}-50{M}:
  Pushing the limits of paraphrastic sentence embeddings with millions of
  machine translations}.
\newblock In \emph{Proceedings of the 56th Annual Meeting of the Association
  for Computational Linguistics (Volume 1: Long Papers)}, pages 451--462,
  Melbourne, Australia. Association for Computational Linguistics.

\bibitem[{Williams et~al.(2018)Williams, Nangia, and Bowman}]{N18-1101}
Adina Williams, Nikita Nangia, and Samuel Bowman. 2018.
\newblock \href {http://aclweb.org/anthology/N18-1101} {A broad-coverage
  challenge corpus for sentence understanding through inference}.
\newblock In \emph{Proceedings of the 2018 Conference of the North American
  Chapter of the Association for Computational Linguistics: Human Language
  Technologies, Volume 1 (Long Papers)}. Association for Computational
  Linguistics.

\bibitem[{Wolf et~al.(2019)Wolf, Debut, Sanh, Chaumond, Delangue, Moi, Cistac,
  Rault, Louf, Funtowicz, and Brew}]{Wolf2019HuggingFacesTS}
Thomas Wolf, Lysandre Debut, Victor Sanh, Julien Chaumond, Clement Delangue,
  Anthony Moi, Pierric Cistac, Tim Rault, R'emi Louf, Morgan Funtowicz, and
  Jamie Brew. 2019.
\newblock Huggingface's transformers: State-of-the-art natural language
  processing.
\newblock \emph{ArXiv}, abs/1910.03771.

\bibitem[{Yu et~al.(2018)Yu, Dohan, Le, Luong, Zhao, and Chen}]{wei2018fast}
Adams~Wei Yu, David Dohan, Quoc Le, Thang Luong, Rui Zhao, and Kai Chen. 2018.
\newblock {QANet: Combining Local Convolution with Global Self-Attention for
  Reading Comprehension}.
\newblock In \emph{International Conference on Learning Representations}.

\bibitem[{Zhang and Bansal(2019)}]{zhang2019addressing}
Shiyue Zhang and Mohit Bansal. 2019.
\newblock {Addressing Semantic Drift in Question Generation for Semi-Supervised
  Question Answering}.
\newblock In \emph{Empirical Methods in Natural Language Processing (EMNLP)}.

\bibitem[{Zhang et~al.(2020)Zhang, Kishore, Wu, Weinberger, and
  Artzi}]{Zhang2020BERTScore}
Tianyi Zhang, Varsha Kishore, Felix Wu, Kilian~Q. Weinberger, and Yoav Artzi.
  2020.
\newblock \href {https://openreview.net/forum?id=SkeHuCVFDr} {{BERTScore:
  Evaluating Text Generation with BERT}}.
\newblock In \emph{International Conference on Learning Representations}.

\bibitem[{Zhang et~al.(2019)Zhang, Baldridge, and He}]{zhang2019paws}
Yuan Zhang, Jason Baldridge, and Luheng He. 2019.
\newblock {PAWS: Paraphrase Adversaries from Word Scrambling}.
\newblock In \emph{Proceedings of the 2019 Conference of the North American
  Chapter of the Association for Computational Linguistics: Human Language
  Technologies, Volume 1 (Long and Short Papers)}.

\bibitem[{Zhou and Bansal(2020)}]{zhou2020towards}
Xiang Zhou and Mohit Bansal. 2020.
\newblock {Towards Robustifying NLI Models Against Lexical Dataset Biases}.
\newblock In \emph{Proceedings of the 58th Annual Meeting of the Association
  for Computational Linguistics}.

\end{thebibliography}
\bibliographystyle{acl_natbib}

\appendix

\section{Dependency Label Set}
As outlined in Section 2, we restrict our analysis to a subset of dependency arcs which are more strongly connected to semantics. For each of hypothesis $h$ corresponding to the input $x$, we extract Enhanced Dependencies using the Stanford CoreNLP tool, and assign entailment labels to this dependency arc set $d(h)$ using the strategy outlined in Section 4. We exclude the following arcs from our analysis: \emph{punct, det, case, aux, auxpass, dep, cop, mark}. This same subset of arcs are ignored while computing sentence-level factuality. 



\section{Examples from Synonym Test Set}
\begin{figure}[h]
\centering
    \includegraphics[trim=15mm 170mm 120mm 15mm,scale=0.44,clip]{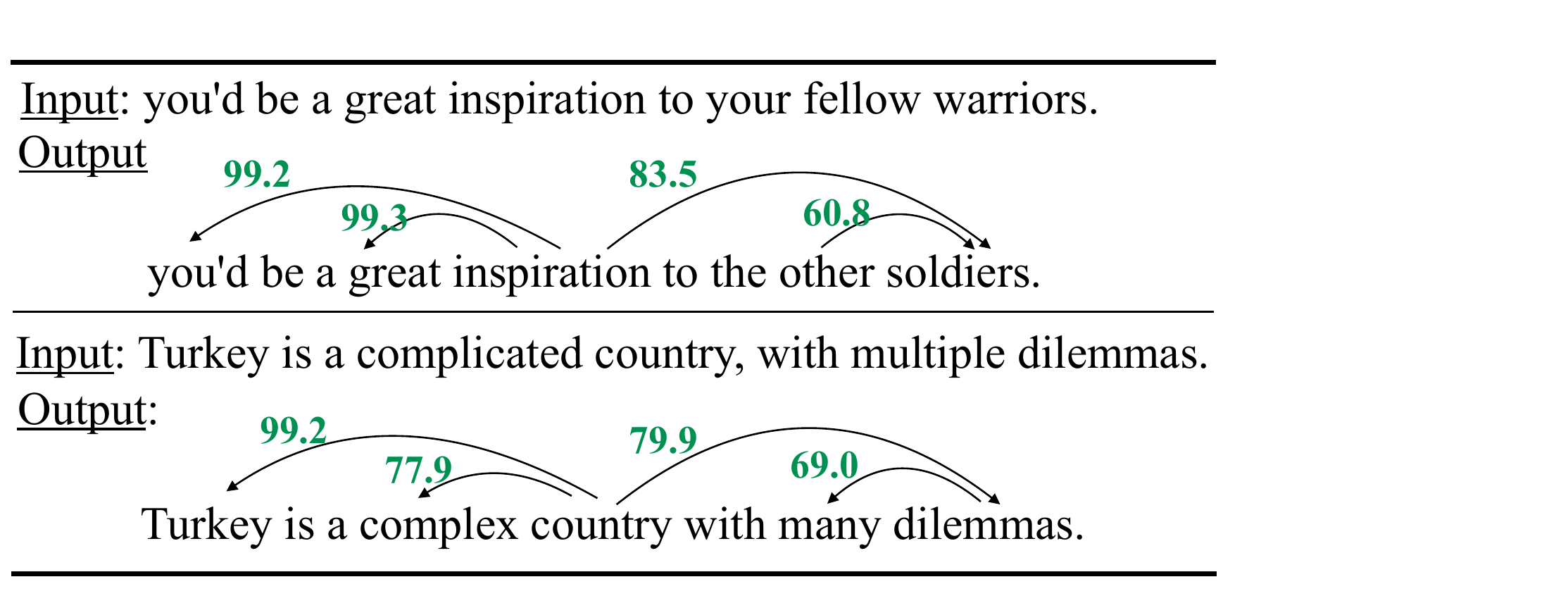}
    \caption{Example from the held-out synonym dataset. The scores are the arc entailment probabilities assigned by the (AD + S) model.} 
    \label{fig:example-syn}
\end{figure}

As outlined in Section 5.2, we additionally augment our auto-derived training data (AD) with synonym data (S) and show that this improves the model performance on the held out synonym only test set. Figure~\ref{fig:example-syn} provides some examples showing the predicted entailment probability for each arc using this augmented training data. The predictions show that our model learns some bias to recognize synonym replacements and small phrasal substitutions as arcs that are entailed by the input.

\section{Implementation Details}
\label{appendix:implementaion}
To train our DAE model, we fine-tune the pre-trained encoders \textsc{Bert} (bert-base-uncased, 110M parameters) and \textsc{Electra} (electra-base-discriminator, 110M parameters), as outlined in Section 5. We perform $5$ hyperparameter trials, varying only the learning rate using manual tuning. The models with the best dev set accuracy are used. The final hyperparameters used are: 

\begin{table}[h]
    \small
    \begin{tabular}{l|l}\toprule
        Implementation Library & transformers \cite{Wolf2019HuggingFacesTS} \\
        Computing Infrastructure & 32GB NVIDIA V100 GPU \\
        Max Seq Length & 128 \\
        Linear Layer Size & (2304, 2)\\
        Optimizer & Adam\\
        Optimizer Params & $\beta=(0.9, 0.999), \epsilon=10^{-8}$ \\
        Learning Rate Decay & Linear \\
        Learning rate & 1e-5\\
        Weight Decay & 0 \\
        Warmup Steps & 0 \\
        Maximum Gradient Norm & 1 \\
        Batch size & 32 \\
        Epochs & 3 \\
         \bottomrule
    \end{tabular}
    \caption{Hyperparameters used for fine-tuning both the \textsc{Bert} and \textsc{Electra} based DAE models.}
    \label{table:params-sow}
\end{table}

Additionally, we fine-tune (bert-base-uncased, 110M parameters) and \textsc{Electra} (electra-base-discriminator, 110M parameters) models on the MNLI dataset. We fine-tuned the model using $3$ hyperparameter trials, varying only the learning rate using manual tuning. The models with the best dev set accuracy are used. The final hyperparameters used are shown in Table \ref{table:params-mnli}.

\begin{table}[h]
    \small
    \begin{tabular}{l|l}\toprule
        Implementation Library & transformers \cite{Wolf2019HuggingFacesTS} \\
        Computing Infrastructure & 32GB NVIDIA V100 GPU \\
        Max Seq Length & 256 \\
        Linear Layer Size & (768, 2)\\
        Optimizer & Adam\\
        Optimizer Params & $\beta=(0.9, 0.999), \epsilon=10^{-8}$ \\
        Learning Rate Decay & Linear \\
        Learning rate & 2e-5\\
        Weight Decay & 0 \\
        Warmup Steps & 0 \\
        Maximum Gradient Norm & 1 \\
        Batch size & 32 \\
        Epochs & 3 \\
         \bottomrule
    \end{tabular}
    \caption{Hyperparameters used for fine-tuning both the \textsc{Bert} and \textsc{Electra} based entailment models.}
    \label{table:params-mnli}
\end{table}

We get a dev accuracy of $84.5\%$ and $89.0\%$ for the \textsc{Bert} and \textsc{Electra} models respectively.  

\end{document}